# Input Parameters Optimization in Swarm DS-CDMA Multiuser Detectors

Taufik Abrão[1], Leonardo D. Oliveira[2], Bruno A. Angélico[3], Paul Jean E. Jeszensky[2]

[1]Dept. of Electrical Engineering; State University of Londrina, Brazil; `taufik@uel.br`
`http://www.uel.br/pessoal/taufik`   `www.uel.br/pessoal/taufik`

[2]Dept. of Telecomm. and Control Engineering, Escola Politécnica of the University of São Paulo, Brazil;
`{leonardo, pjj}@lcs.poli.usp.br`

[3]Federal University of Technology - Paraná (UTFPR), Cornélio Procópio, Brazil; `bangelico@utfpr.edu.br`

**Abstract**

In this paper, the uplink direct sequence code division multiple access (DS-CDMA) multiuser detection problem (MUD) is studied into heuristic perspective, named particle swarm optimization (PSO). Regarding different system improvements for future technologies, such as high-order modulation and diversity exploitation, a complete parameter optimization procedure for the PSO applied to MUD problem is provided, which represents the major contribution of this paper. Furthermore, the performance of the PSO-MUD is briefly analyzed via Monte-Carlo simulations. Simulation results show that, after convergence, the performance reached by the PSO-MUD is much better than the conventional detector, and somewhat close to the single user bound (SuB). Rayleigh flat channel is initially considered, but the results are further extend to diversity (time and spatial) channels.

## I. INTRODUCTION

In a DS-CDMA system, a conventional detector by itself may not provide a desirable quality of service, once the system capacity is strongly affected by multiple access interference (MAI). The capacity of a DS-CDMA system in multipath channels is limited mainly by the MAI, self-interference (SI), near-far effect (NFR) and fading. The conventional receiver (Rake) explores the path diversity in order to reduce fading effect, but it is not able to mitigate neither the MAI nor the near-far effect [14], [25]. In this context, multiuser detection emerged as a solution to overcome the MAI [25]. The best performance is acquired by the optimum multiuser detection (OMuD), based on the log-likelihood function (LLF) [25]. In [24] it was demonstrated that multiuser detection problem results in a nondeterministic polynomial-time hard (NP-hard) problem. After the Verdu's revolutionary work, a great variety of suboptimal approaches have been proposed: from linear multiuser detectors [25], [2] to heuristic multiuser detectors [9], [7].

Alternatives to OMuD into the class of linear multiuser detectors include the Decorrelator [23], and MMSE [19]. Besides, the classic non-linear multiuser detectors include the subtractive interference cancellation (IC) MuD [18] and zero-forcing decision feedback (ZFDF) [5]. In spite of the relatively low complexity, the drawback of (non-)linear, ZFDF, and hybrid cancelers sub-optimal MuDs is the failure in approaching the ML performance under realistic channel and system scenarios. More recently heuristic methods have been proposed for solving the MuD problem, obtaining near-ML performance at cost of polynomial computational complexity [7], [1]. Examples of heuristic multiuser detection (HEUR-MuD) methods include: evolutionary programming (EP), specially the genetic algorithm (GA) [7], [4], particle swarm optimization (PSO) [12], [26], [16] and, sometimes included in this classification, the deterministic local search (LS) methods [13], [17], which has been shown to present an very attractive performance $\times$ complexity trade-off for low order modulations.

Nevertheless, there are few works dealing with complex and realistic system configurations. High-order modulation HEUR-MuD in SISO or MIMO systems were previously addressed in [12], [26], [15]. In [15], PSO was applied to near-optimum asynchronous DS-CDMA multiuser detection problem under $16-$QAM modulation and SISO multipath channels. Previous results on literature [16], [1] suggest that evolutionary algorithms and particle swarm optimization have similar performance, and that a simple local search heuristic optimization is enough to solve the MuD problem with low-order modulation [17]. However for high-order modulation formats, the LS-MuD does not achieve good performances due to a lack of search diversity, whereas the PSO-MuD has been shown to be more efficient for solving the optimization problem under $M$-QAM modulation [15].

Recent works applying PSO to MuD usually assumes conventional values for PSO input parameters, such [10], or optimized values only for specific system and channel scenarios, such [16] for flat Rayleigh channel, [15] for multipath and high-order modulation, and [1] for multicarrier CDMA systems as well. In this paper, a wide analysis, with BPSK, QPSK and 16-QAM modulation schemes, and diversity exploration is carried out.

This paper provides a quite complete parameter optimization of the PSO-MuD applied to DS-CDMA systems in Rayleigh channels with BPSK, QPSK and 16-QAM modulations. The text has the following organization: Section II presents the system model, including DS-CDMA, OMuD, and the PSO-MuD. The PSO parameter optimization is shown in Section III, while Section IV exhibits some performance results in terms of Monte Carlo simulation (MCS). Finally, Section V summarizes the main conclusions of this work.

## II. SYSTEM MODEL

In this Section, a single-cell asynchronous multiple access DS-CDMA system model is described for Rayleigh channels, considering different modulation schemes, such as binary/quadrature phase shift keying (BPSK/QPSK) and 16-quadrature amplitude modulation (16-QAM), and single or multiple antennas at the base station receiver. After describing the conventional detection approach with a maximum ratio combining (MRC) rule, the OMUD and the PSO-MUD are described. The model is generic enough to allow describing additive white Gaussian noise (AWGN) and Rayleigh flat channels, other modulation formats and single-antenna receiver.

### A. DS-CDMA

The base-band transmitted signal of the $k$th user is described as [20]

$$s_k(t) = \sqrt{\frac{\mathcal{E}_k}{T}} \sum_{i=-\infty}^{\infty} d_k^{(i)} g_k(t - iT), \tag{1}$$

where $\mathcal{E}_k$ is the symbol energy, and $T$ is the symbol duration. Each symbol $d_k^{(i)}$, $k = 1, \ldots, K$ is taken independently and with equal probability from a complex alphabet set $\mathcal{A}$ of cardinality $M = 2^m$ in a squared constellation, i.e., $d_k^{(i)} \in \mathcal{A} \subset \mathbb{C}$, where $\mathbb{C}$ is the set of complex numbers. Fig. 1 shows the modulation formats considered, while Fig. 2 sketches the $K$ base-band DS-CDMA transmitters.

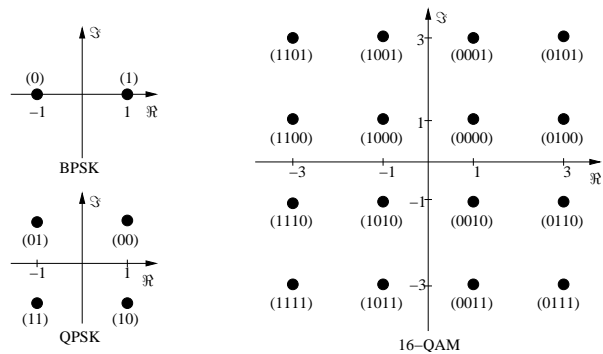

Fig. 1. Three modulation formats with Gray mapping.

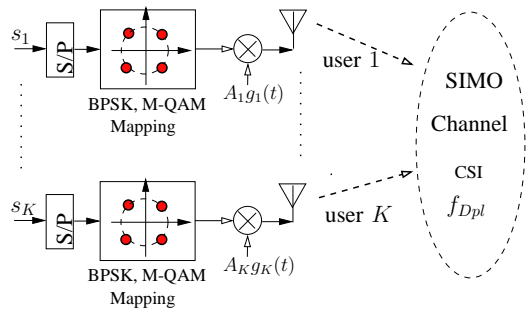

Fig. 2. Uplink base-band DS-CDMA transmission model with $K$ users.

The normalized spreading sequence for the $k$-th user is given by

$$g_k(t) = \frac{1}{\sqrt{N}} \sum_{n=0}^{N-1} a_k(n) p(t - nT_c), \qquad 0 \leq t \leq T, \tag{2}$$

where $a_k(n)$ is a random sequence with $N$ chips assuming the values $\{\pm 1\}$, $p(t)$ is the pulse shaping, assumed rectangular with unitary amplitude and duration $T_c$, with $T_c$ being the chip interval. The processing gain is given by $N = T/T_c$.

The equivalent base-band received signal at $q$th receive antenna, $q = 1, 2, \ldots, Q$, containing $I$ symbols for each user in multipath fading channel can be expressed by

$$\begin{aligned} r_q(t) &= \sum_{i=0}^{I-1} \sum_{k=1}^{K} \sum_{\ell=1}^{L} A_k d_k^{(i)} g_k(t - nT - \tau_{q,k,\ell}) \\ &\quad h_{q,k,\ell}^{(i)} e^{j\varphi_{q,k,\ell}} + \eta_q(t), \end{aligned} \tag{3}$$

with $A_k = \sqrt{\frac{\mathcal{E}_k}{T}}$, $L$ being the number of channel paths, admitted equal for all $K$ users, $\tau_{q,k,\ell}$ is the total delay[1] for the signal of the $k$th user, $\ell$th path at $q$th receive antenna, $e^{j\varphi_{q,k,\ell}}$ is the respective received phase carrier; $\eta_q(t)$ is the additive white Gaussian noise with bilateral power spectral density equal to $N_0/2$, and $h_{q,k,\ell}^{(i)}$ is the complex channel coefficient for the $i$th symbol, defined as

$$h_{q,k,\ell}^{(i)} = \gamma_{q,k,\ell}^{(i)} e^{j\theta_{q,k,\ell}^{(i)}}, \tag{4}$$

where the gain $\gamma_{q,k,\ell}^{(i)}$ is a characterized by a Rayleigh distribution and the phase $\theta_{q,k,\ell}^{(i)}$ by the uniform distribution $\mathcal{U}[0, 2\pi]$.

Generally, a slow and frequency selective channel[2] is assumed. The expression in (3) is quite general and includes some special and important cases: if $Q = 1$, a SISO system is obtained; if $L = 1$, the channel becomes non-selective (flat) Rayleigh; if $h_{q,k,\ell}^{(i)} = 1$, it results in the AWGN channel; moreover, if $\tau_{q,k,\ell} = 0$, a synchronous DS-CDMA system is characterized.

At the base station, the received signal is submitted to a matched filter bank (CD), with $D \leq L$ branches (fingers) per antenna of each user. When $D \geq 1$, CD is known as Rake receiver. Assuming perfect phase estimation (carrier phase), after despreading the resultant signal is given by

$$\begin{aligned} y_{q,k,\ell}^{(i)} &= \frac{1}{T} \int_{nT}^{(i+1)T} r_q(t) g_k(t - \tau_{q,k,\ell}) dt \\ &= A_k h_{q,k,\ell}^{(i)} d_k^{(i)} + SI_{q,k,\ell}^{(i)} + I_{q,k,\ell}^{(i)} + \widetilde{\eta}_{q,k,\ell}^{(i)}. \end{aligned} \tag{5}$$

The first term is the signal of interest, the second corresponds to the self-interference (SI), the third to the multiple-access interference (MAI) and the last one corresponds to the filtered AWGN.

---

[1] Considering the asynchronism among the users and random delays for different paths.

[2] Slow channel: channel coefficients were admitted constant along the symbol period $T$; and frequency selective condition is hold: $\frac{1}{T_c} \gg (\Delta B)_c$, the coherence bandwidth of the channel.

Considering a maximal ratio combining (MRC) rule with diversity order equal to $DQ$ for each user, the $M-$level complex decision variable is given by

$$\zeta_k^{(i)} = \sum_{q=1}^{Q}\sum_{\ell=1}^{D} y_{q,k,\ell}^{(i)} \cdot \mathtt{w}_{q,k,\ell}^{(i)}, \quad k=1,\ldots,K \qquad (6)$$

where the MRC weights $\mathtt{w}_{q,k,\ell}^{(i)} = \widehat{\gamma}_{q,k,\ell}^{(i)} e^{-j\widehat{\theta}_{q,k,\ell}^{(i)}}$, with $\widehat{\gamma}_{q,k,l}^{(i)}$ and $\widehat{\theta}_{q,k,\ell}^{(i)}$ been a channel amplitude and phase estimation, respectively.

After that, at each symbol interval, decisions are made on the in-phase and quadrature components[3] of $\zeta_k^{(i)}$ by scaling it into the constellation limits obtaining $\xi_k^{(i)}$, and choosing the complex symbol with minimum Euclidean distance regarding the scaled decision variable. Alternatively, this procedure can be replaced by separate $\sqrt{M}-$level quantizers $\mathtt{qtz}$ acting on the in-phase and quadrature terms separately, such that

$$\widehat{d}_k^{(i),\text{CD}} = \underset{\mathcal{A}_{\text{real}}}{\mathtt{qtz}}\left(\Re\left\{\xi_k^{(i)}\right\}\right) + j\underset{\mathcal{A}_{\text{imag}}}{\mathtt{qtz}}\left(\Im\left\{\xi_k^{(i)}\right\}\right), \qquad (7)$$

for $k = 1,\ldots,K$, and where $\mathcal{A}_{\text{real}}$ and $\mathcal{A}_{\text{imag}}$ is the real and imaginary value sets, respectively, from the complex alphabet set $\mathcal{A}$, and $\Re\{\cdot\}$ and $\Im\{\cdot\}$ representing the real and imaginary operators, respectively. Fig. 3 illustrates the general system structure.

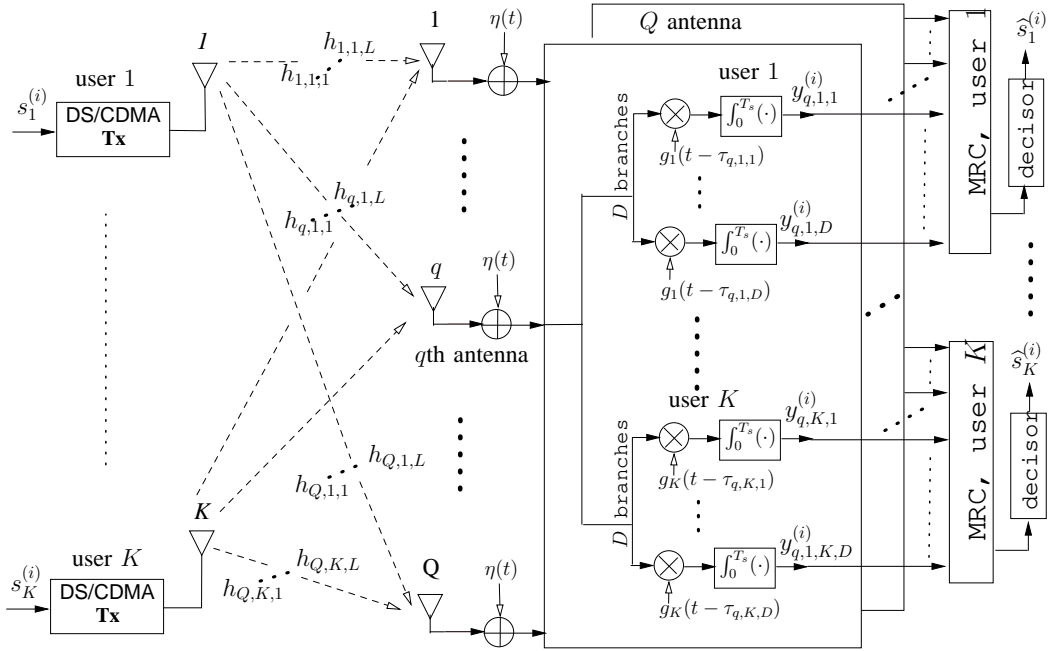

Fig. 3. Uplink base-band DS-CDMA system model with Conventional receiver: $K$ users transmitters, SIMO channel and conventional (Rake) receiver with $Q$ multiple receive antennas.

---

[3]Note that, for BPSK, only the in-phase term is presented.

*B. Optimum Detection*

The OMUD estimates the symbols for all $K$ users by choosing the symbol combination associated with the minimal distance metric among all possible symbol combinations in the $M = 2^m$ constellation points [25].

In the asynchronous multipath channel scenario considered in this paper, the one-shot asynchronous channel approach is adopted, where a configuration with $K$ asynchronous users, $I$ symbols and $D$ branches is equivalent to a synchronous scenario with $KID$ virtual users.

Furthermore, in order to avoid handling complex-valued variables in high-order squared modulation formats, henceforward the alphabet set is re-arranged as $\mathcal{A}_{\text{real}} = \mathcal{A}_{\text{imag}} = \mathcal{Y} \subset \mathbb{Z}$ of cardinality $\sqrt{M}$, e.g., 16$-$QAM ($m = 4$): $d_k^{(i)} \in \mathcal{Y} = \{\pm 1, \pm 3\}$.

The OMUD is based on the maximum likelihood criterion that chooses the vector of symbols $\underline{\mathbf{d}}_p$, formally defined in (12), which maximizes the metric

$$\underline{\mathbf{d}}^{\text{opt}} = \arg \max_{\underline{\mathbf{d}}_p \in \mathcal{Y}^{2KID}} \{\Omega(\underline{\mathbf{d}}_p)\}, \qquad (8)$$

where, in a SIMO channel, the single-objective function is generally written as a combination of the LLFs from all receive antennas, given by

$$\Omega(\underline{\mathbf{d}}_p) = \sum_{q=1}^{Q} \Omega_q(\underline{\mathbf{d}}_p). \qquad (9)$$

In the more general case considered here, i.e., $K$ asynchronous users in a SIMO multipath Rayleigh channel with diversity $D \leq L$, the LLF can be defined as a decoupled optimization problem with only real-valued variables, such that

$$\Omega_q(\underline{\mathbf{d}}_p) = 2\underline{\mathbf{d}}_p^\top \mathbf{W}_q^\top \underline{\mathbf{y}}_q - \underline{\mathbf{d}}_p^\top \mathbf{W}_q \underline{\mathbf{R}} \mathbf{W}_q^\top \underline{\mathbf{d}}_p, \qquad (10)$$

with definitions

$$\underline{\mathbf{y}}_q := \begin{bmatrix} \Re\{\mathbf{y}_q\} \\ \Im\{\mathbf{y}_q\} \end{bmatrix}; \quad \mathbf{W}_q := \begin{bmatrix} \Re\{\mathbf{AH}\} & -\Im\{\mathbf{AH}\} \\ \Im\{\mathbf{AH}\} & \Re\{\mathbf{AH}\} \end{bmatrix};$$

$$\underline{\mathbf{d}}_p := \begin{bmatrix} \Re\{\mathbf{d}_p\} \\ \Im\{\mathbf{d}_p\} \end{bmatrix}; \quad \underline{\mathbf{R}} := \begin{bmatrix} \mathbf{R} & \mathbf{0} \\ \mathbf{0} & \mathbf{R} \end{bmatrix}, \qquad (11)$$

where $\underline{\mathbf{y}}_q \in \mathbb{R}^{2KID \times 1}$, $\mathbf{W}_q \in \mathbb{R}^{2KID \times 2KID}$, $\underline{\mathbf{d}}_p \in \mathcal{Y}^{2KID \times 1}$, $\underline{\mathbf{R}} \in \mathbb{R}^{2KID \times 2KID}$. The vector $\mathbf{d}_p \in \mathcal{Y}^{KID \times 1}$ in Eq. (11) is defined as

$$\mathbf{d}_p = [(\underbrace{d_1^{(1)} \cdots d_1^{(1)}}_{D\,\text{times}}) \cdots (\underbrace{d_K^{(1)} \cdots d_K^{(1)}}_{D\,\text{times}}) \cdots (\underbrace{d_1^{(I)} \cdots d_1^{(I)}}_{D\,\text{times}}) \cdots (\underbrace{d_K^{(I)} \cdots d_K^{(I)}}_{D\,\text{times}})]^\top. \qquad (12)$$

In addition, the $\mathbf{y}_q \in \mathbb{C}^{KID \times 1}$ is the despread signal in Eq. (6) for a given $q$, in a vector notation, described as

$$\underline{\mathbf{y}}_q = \left[ (y_{q,1,1}^{(1)} \cdots y_{q,1,D}^{(1)}) \cdots (y_{q,K,1}^{(1)} \cdots y_{q,K,D}^{(1)}) \cdots \right.$$
$$\left. (y_{q,1,1}^{(I)} \cdots y_{q,1,D}^{(I)}) \cdots (y_{q,K,1}^{(I)} \cdots y_{q,K,D}^{(I)}) \right] \tag{13}$$

Matrices $\mathbf{H}$ and $\mathbf{A}$ are the coefficients and amplitudes diagonal matrices, and $\mathbf{R}$ represents the block-tridiagonal, block-Toeplitz cross-correlation matrix, composed by the sub-matrices $\mathbf{R}[1]$ and $\mathbf{R}[0]$, such that [25]

$$\mathbf{R} = \begin{bmatrix} \mathbf{R}[0] & \mathbf{R}[1]^\top & \mathbf{0} & \cdots & \mathbf{0} & \mathbf{0} \\ \mathbf{R}[1] & \mathbf{R}[0] & \mathbf{R}[1]^\top & \cdots & \mathbf{0} & \mathbf{0} \\ \mathbf{0} & \mathbf{R}[1] & \mathbf{R}[0] & \cdots & \mathbf{0} & \mathbf{0} \\ \cdots & \cdots & \cdots & \cdots & \cdots & \cdots \\ \mathbf{0} & \mathbf{0} & \mathbf{0} & \cdots & \mathbf{R}[1] & \mathbf{R}[0] \end{bmatrix}, \tag{14}$$

with $\mathbf{R}[0]$ and $\mathbf{R}[1]$ being $KD$ matrices with elements

$$\underline{\rho_{a,b}}[0] = \begin{cases} 1, & \text{if } (k = u) \text{ and } (\ell = l) \\ \rho_{k,\ell,u,l}^q, & \text{if } (k < u) \text{ or } (k = u, \ell < l) \\ \rho_{u,l,k,\ell}^q, & \text{if } (k > u) \text{ or } (k = u, \ell > l), \end{cases}$$

$$\underline{\rho_{a,b}}[1] = \begin{cases} 0, & \text{if } k \geq u \\ \rho_{u,l,k,\ell}^q, & \text{if } k < u \end{cases}, \tag{15}$$

where $a = (k-1)D + \ell$, $b = (u-1)D + l$ and $k, u = 1, 2, \ldots, K$; $\ell, l = 1, 2, \ldots, D$; the cross-correlation element between the $k$th user, $\ell$th path and $u$th user, $d$th path, at $q$th receive antenna, $\rho_{k,\ell,u,d}^q$, is

$$\rho_{k,\ell,u,d}^q = \frac{1}{T} \int_0^T g_k(t - \tau_{q,k,\ell}) g_u(t - \tau_{q,u,d}) dt. \tag{16}$$

The evaluation in (8) can be extended along the whole message, where all symbols of the transmitted vector for all $K$ users are jointly detected (vector ML approach), or the decisions can be taken considering the optimal single symbol detection of all $K$ multiuser signals (symbol ML approach). In the synchronous case, the symbol ML approach with $I = 1$ is considered, whereas in the asynchronous case the vector ML approach is adopted with $I = 7$ ($I$ must be, at least, equal to three ($I \geq 3$)).

The vector $\underline{\mathbf{d}}_p$ in (11) belongs to a discrete set with size depending on $M$, $K$, $I$ and $D$. Hence, the optimization problem posed by (8) can be solved directly using a $m-$dimensional ($m = \log_2 M$) search method. Therefore, the associated combinatorial problem strictly requires an exhaustive search in $\mathcal{A}^{KID}$ possibilities of $\mathbf{d}$, or equivalently an exhaustive search in $\mathcal{Y}^{2KID}$ possibilities of $\underline{\mathbf{d}}_p$ for the decoupled optimization problem with only real-valued variables. As a result, the maximum likelihood detector has a complexity that increases exponentially with the modulation order, number of users, symbols and branches, becoming prohibitive even for moderate product values $mKID$, i.e., even for a BPSK modulation format, medium system loading $(K/N)$, small number of symbols $I$ and $D$ Rake fingers.

## C. Discrete Swarm Optimization Algorithm

A discrete or, in several cases, binary PSO [11] is considered in this paper. Such scheme is suitable to deal with digital information detection/decoding. Hence, binary PSO is adopted herein. The particle selection for evolving is based on the highest fitness values obtained through (10) and (9).

Accordingly, each candidate-vector defined like $\mathbf{d}_i$ has its binary representation, $\mathbf{b}_p[\mathtt{t}]$, of size $mKI$, used for the velocity calculation, and the $p$th PSO particle position at instant (iteration) $\mathtt{t}$ is represented by the $mKI \times 1$ binary vector

$$\begin{aligned} \mathbf{b}_p[\mathtt{t}] &= [\mathbf{b}_p^1\, \mathbf{b}_p^2\, \cdots\, \mathbf{b}_p^r\, \cdots\, \mathbf{b}_p^{KI}]; \qquad (17)\\ \mathbf{b}_p^r &= \left[b_{p,1}^r \cdots b_{p,\nu}^r \cdots b_{p,m}^r\right];\quad b_{p,\nu}^r \in \{0,\, 1\}, \end{aligned}$$

where each binary vector $\mathbf{b}_p^r$ is associated with one $d_k^{(i)}$ symbol in Eq. (12). Each particle has a velocity, which is calculated and updated according to

$$\mathbf{v}_p[\mathtt{t}+1] = \omega \cdot \mathbf{v}_p[\mathtt{t}] + \phi_1 \cdot \mathbf{U}_{p_1}[\mathtt{t}](\mathbf{b}_p^{\text{best}}[\mathtt{t}] - \mathbf{b}_p[\mathtt{t}]) + \phi_2 \cdot \mathbf{U}_{p_2}[\mathtt{t}](\mathbf{b}_g^{\text{best}}[\mathtt{t}] - \mathbf{b}_p[\mathtt{t}]), \qquad (18)$$

where $\omega$ is the inertial weight; $\mathbf{U}_{p_1}[\mathtt{t}]$ and $\mathbf{U}_{p_2}[\mathtt{t}]$ are diagonal matrices with dimension $mKI$, whose elements are random variables with uniform distribution $\mathcal{U} \in [0, 1]$; $\mathbf{b}_g^{\text{best}}[\mathtt{t}]$ and $\mathbf{b}_p^{\text{best}}[\mathtt{t}]$ are the best global position and the best local positions found until the $\mathtt{t}$th iteration, respectively; $\phi_1$ and $\phi_2$ are weight factors (acceleration coefficients) regarding the best individual and the best global positions influences in the velocity update, respectively.

For MUD optimization with binary representation, each element in $\mathbf{b}_p[\mathtt{t}]$ in (18) just assumes "0" or "1" values. Hence, a discrete mode for the position choice is carried out inserting a probabilistic decision step based on threshold, depending on the velocity. Several functions have this characteristic, such as the sigmoid function [11]

$$S(v_{p,\nu}^r[\mathtt{t}]) = \frac{1}{1+e^{-v_{p,\nu}^r[\mathtt{t}]}}, \qquad (19)$$

where $v_{p,\nu}^r[\mathtt{t}]$ is the $r$th element of the $p$th particle velocity vector, $\mathbf{v}_p^r = \left[v_{p,1}^r \cdots v_{p,\nu}^r \cdots v_{p,m}^r\right]$, and the selection of the future particle position is obtained through the statement

$$\begin{aligned} \text{if} \quad \mathtt{u}_{p,\nu}^r[\mathtt{t}] &< S(v_{p,\nu}^r[\mathtt{t}]), \quad b_{p,\nu}^r[\mathtt{t}+1] = 1;\\ \text{otherwise}, &\qquad\qquad\qquad b_{p,\nu}^r[\mathtt{t}+1] = 0, \end{aligned} \qquad (20)$$

where $b_{p,\nu}^r[\mathrm{t}]$ is an element of $\mathbf{b}_p[\mathrm{t}]$ (see Eq. (18)), and $\mathrm{u}_{p,\nu}^r[\mathrm{t}]$ is a random variable with uniform distribution $\mathcal{U} \in [0, 1]$.

After obtaining a new particle position $\mathbf{b}_p[\mathrm{t}+1]$, it is mapped back into its correspondent symbol vector $\mathbf{d}_p[\mathrm{t}+1]$, and further in the real form $\underline{\mathbf{d}}_p[\mathrm{t}+1]$, for the evaluation of the objective function in (9).

In order to obtain further diversity for the search universe, the $V_{\max}$ factor is added to the PSO model, Eq. (18), being responsible for limiting the velocity in the range $[\pm V_{\max}]$. The insertion of this factor in the velocity calculation enables the algorithm to escape from possible local optima. The likelihood of a bit change increases as the particle velocity crosses the limits established by $[\pm V_{\max}]$, as shown in Tab. I.

TABLE I

MINIMUM BIT CHANGE PROBABILITY AS A FUNCTION OF $V_{\max}$.

| $V_{\max}$ | 1 | 2 | 3 | 4 | 5 |
|---|---|---|---|---|---|
| $1 - S(V_{\max})$ | 0.269 | 0.119 | 0.047 | 0.018 | 0.007 |

Population size $\mathcal{P}$ is typically in the range of 10 to 40 [6]. However, based on [16], it is set to

$$\mathcal{P} = 10 \left\lfloor 0.3454 \left( \sqrt{\pi(mKI - 1)} + 2 \right) \right\rfloor. \tag{21}$$

Algorithm 1 describes the pseudo-code for the PSO implementation.

## III. PSO-MUD PARAMETERS OPTIMIZATION

In this section, the PSO-MUD parameters optimization is carried out using Monte Carlo simulation. Such an optimization is directly related to the complexity $\times$ performance trade-off of the algorithm. A wide analysis with BPSK, QPSK and 16-QAM modulation schemes, and diversity exploration is carried out.

A first analysis of the PSO parameters gives raise to the following behaviors: $\omega$ is responsible for creating an inertia of the particles, inducing them to keep the movement towards the last directions of their velocities; $\phi_1$ aims to guide the particles to each individual best position, inserting diversification in the search; $\phi_2$ leads all particles towards the best global position, hence intensifying the search and reducing the convergence time; $V_{\max}$ inserts perturbation limits in the movement of the particles, allowing more or less diversification in the algorithm.

The optimization process for the initial velocity of the particles achieves similar results for three different conditions: null, random and CD output as initial velocity. Hence, it is adopted here, for simplicity, null initial velocity, i.e., $\mathbf{v}[0] = \mathbf{0}$.

**Algorithm 1** PSO Algorithm for the MUD Problem
---
**Input:** $\mathbf{d}^{\text{CD}}$, $\mathcal{P}$, $G$, $\omega$, $\phi_1$, $\phi_2$, $V_{\max}$;    **Output:** $\mathbf{d}^{\text{PSO}}$
begin
1.  initialize first population: $\mathtt{t} = 0$;
    $\mathbf{B}[0] = \mathbf{b}^{\text{CD}} \cup \widetilde{\mathbf{B}}$, where $\widetilde{\mathbf{B}}$ contains $(\mathcal{P} - 1)$ particles randomly generated;
    $\mathbf{b}_p^{\text{best}}[0] = \mathbf{b}_p[0]$ and $\mathbf{b}_g^{\text{best}}[0] = \mathbf{b}^{\text{CD}}$;
    $\mathbf{v}_p[0] = \mathbf{0}$: null initial velocity;
2.  while $\mathtt{t} \leq G$
    a. calculate $\Omega(\underline{\mathbf{d}}_p[\mathtt{t}])$, $\forall \mathbf{b}_p[\mathtt{t}] \in \mathbf{B}[\mathtt{t}]$ using (9);
    b. update velocity $\mathbf{v}_p[\mathtt{t}]$, $p = 1, \ldots, \mathcal{P}$, through (18);
    c. update best positions:
       for $p = 1, \ldots, \mathcal{P}$
         if $\Omega(\underline{\mathbf{d}}_p[\mathtt{t}]) > \Omega(\underline{\mathbf{d}}_p^{\text{best}}[\mathtt{t}])$, $\mathbf{b}_p^{\text{best}}[\mathtt{t}+1] \leftarrow \mathbf{b}_p[\mathtt{t}]$
         else $\mathbf{b}_p^{\text{best}}[\mathtt{t}+1] \leftarrow \mathbf{b}_p^{\text{best}}[\mathtt{t}]$
       end
       if $\exists\, \mathbf{b}_p[\mathtt{t}]$ such that $\left[\Omega(\underline{\mathbf{d}}_p[\mathtt{t}]) > \Omega(\underline{\mathbf{d}}_g^{\text{best}}[\mathtt{t}])\right] \wedge$
          $\left[\Omega(\underline{\mathbf{d}}_p[\mathtt{t}]) \geq \Omega(\underline{\mathbf{d}}_j[\mathtt{t}]), j \neq p\right]$,
         $\mathbf{b}_g^{\text{best}}[\mathtt{t}+1] \leftarrow \mathbf{b}_p[\mathtt{t}]$
       else $\mathbf{b}_g^{\text{best}}[\mathtt{t}+1] \leftarrow \mathbf{b}_g^{\text{best}}[\mathtt{t}]$
    d. Evolve to a new swarm population $\mathbf{B}[\mathtt{t}+1]$, using (20);
    e. set $\mathtt{t} = \mathtt{t} + 1$.
    end
3.  $\mathbf{b}^{\text{PSO}} = \mathbf{b}_g^{\text{best}}[G]$;  $\mathbf{b}^{\text{PSO}} \xrightarrow{\text{map}} \mathbf{d}^{\text{PSO}}$.
end
— — — — — — — — — — — — — — — — — — — —
$\mathbf{d}^{\text{CD}}$: CD output.
$\mathcal{P}$: Population size.
$G$: number of swarm iterations.
For each $\underline{\mathbf{d}}_p[\mathtt{t}]$ there is a $\mathbf{b}_p[\mathtt{t}]$ associated.

---

In [16], the best performance $\times$ complexity trade-off for BPSK PSO-MUD algorithm was obtained setting $V_{\max} = 4$. Herein, simulations carried out varying $V_{\max}$ for different modulations and diversity exploration accomplish this value as a good alternative. This optimization process is quite similar for systems with QPSK and 16-QAM modulation formats.

*A. $\omega$ Optimization*

It is worth noting that a relatively larger value for $\omega$ is helpful for global optimum, and lesser influenced by the best global and local positions, while a relatively smaller value for $\omega$ is helpful for course convergence, i.e., smaller inertial weight encourages the local exploration [6], [21] as the particles are more attracted towards $\mathbf{b}_p^{\text{best}}[\mathtt{t}]$ and $\mathbf{b}_g^{\text{best}}[\mathtt{t}]$.

Fig. 4 shows the convergence of the PSO scheme for different values of $\omega$ considering BPSK modulation and flat channel. It is evident that the best performance $\times$ complexity trade-off is accomplished with $\omega = 1$.

Many research papers have been proposed new strategies for PSO principle in order to improve its performance and reduce its complexity. For instance, in [3] the authors have been discussed adaptive nonlinear inertia weight in order to improve PSO convergence. However, the current analysis indicates that no further specialized strategy is necessary, since the conventional PSO works well to solve the MUD DS-CDMA problem in several practical scenarios.

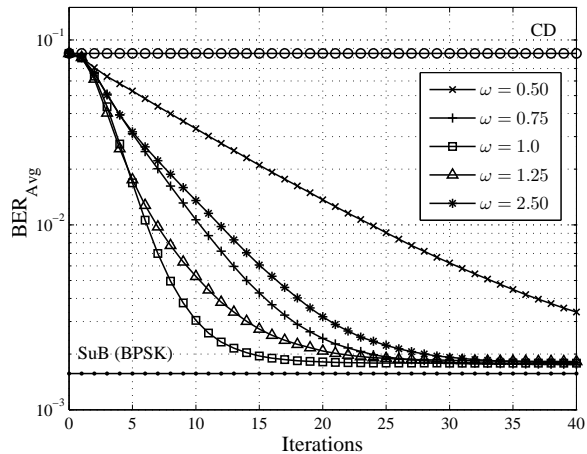

Fig. 4.  $\omega$ optimization under Rayleigh flat channels with BPSK modulation, $E_b/N_0 = 22$ dB, $K = 15$, $\phi_1 = 2$, $\phi_2 = 10$ and $V_{\max} = 4$.

The optimization of the inertial weight, $\omega$, achieves analogous results for QPSK and 16-QAM modulation schemes, where $\omega = 1$ also achieves the best performance $\times$ complexity trade-off (results not shown). A special attention is given for $\phi_1$ and $\phi_2$ optimization in the next, since their values impact deeply in the PSO performance, also varying for each modulation.

## B. $\phi_1$ and $\phi_2$ Optimization

*1) BPSK Modulation:* For Rayleigh channels, the performance improvement expected by $\phi_1$ increment is not evident, and its value can be reduced without performance losses, as can be seen in Fig. 5. Therefore, a good choice seems to be $\phi_1 = 2$, achieving a reasonable convergence rate.

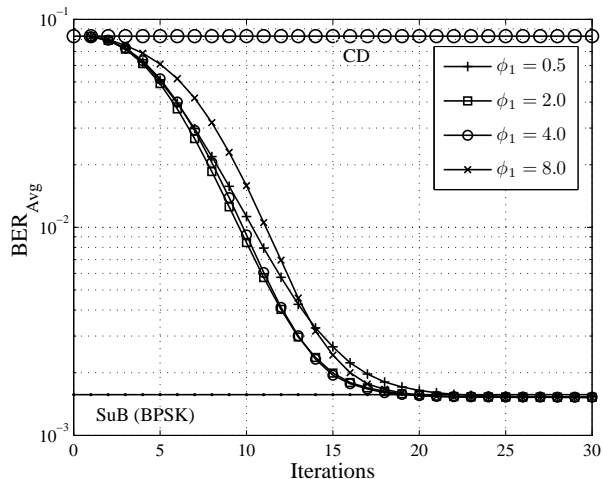

Fig. 5.  $\phi_1$ optimization in Rayleigh flat channels with BPSK modulation, $E_b/N_0 = 22$ dB, $K = 15$, $\phi_2 = 10$, $V_{\max} = 4$, and $\omega = 1$.

Fig. 6.(a) illustrates different convergence performances achieved with $\phi_1 = 2$ and $\phi_2 \in [1; 15]$ for medium system loading and medium-high $E_b/N_0$. Even for high system loading, the PSO performance

is quite similar for different values of $\phi_2$, as observed in Fig. 6.(b). Hence, considering the performance × complexity trade-off, a reasonable choice for $\phi_2$ under Rayleigh flat channels is $\phi_2 = 10$.

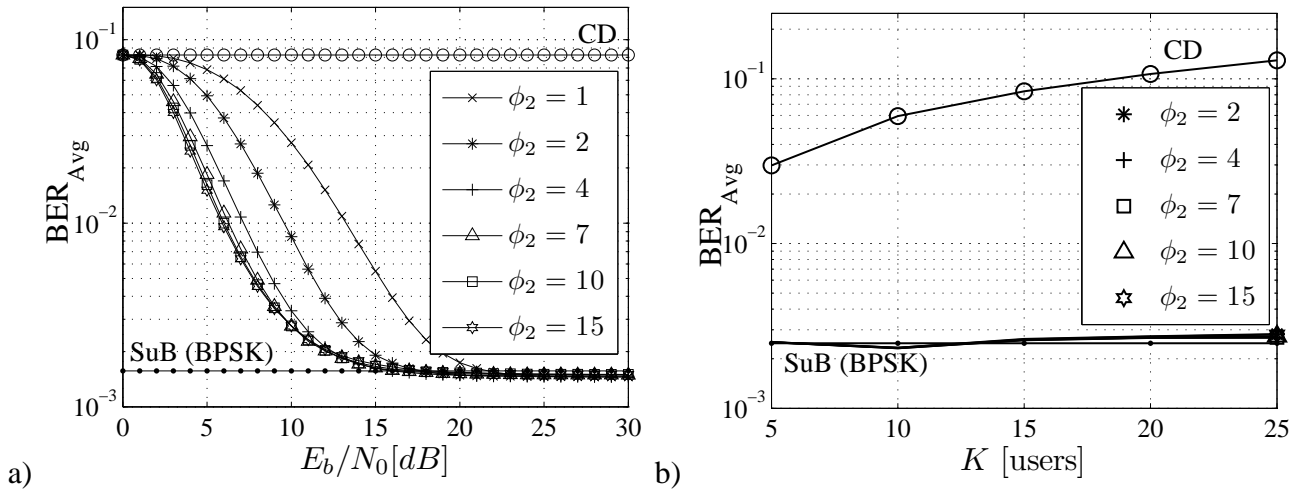

Fig. 6. $\phi_2$ optimization in Rayleigh flat channels with BPSK modulation, $V_{\max} = 4$, $\omega = 1$, and $\phi_1 = 2$; a) convergence performance with $E_b/N_0 = 22$ dB and $K = 15$; b) average BER × $K$ with $E_b/N_0 = 20$ dB, $G = 30$ iterations.

*2) QPSK Modulation:* Different results from BPSK are achieved when a QPSK modulation scheme is adopted. Note in Fig. 7 that low values of $\phi_2$ and high values $\phi_1$ delay the convergence, the inverse results in lack of diversity. Hence, the best performance × complexity is achieved with $\phi_1 = \phi_2 = 4$.

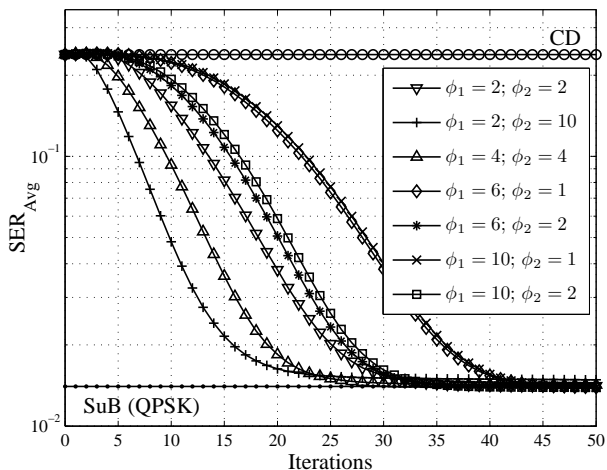

Fig. 7. $\phi_1$ and $\phi_2$ optimization under flat Rayleigh channels for QPSK modulation, $E_b/N_0 = 22$ dB, $K = 15$, $\omega = 1$ and $V_{\max} = 4$.

*3) 16-QAM Modulation:* Under 16-QAM modulation, the PSO-MUD requires more intensification, once the search becomes more complex due to each symbol maps to 4 bits. Fig. 8 shows the convergence curves for different values of $\phi_1$ and $\phi_2$, where it is clear that the performance gap is more evident with an increasing number of users and $E_b/N_0$. Analyzing this result, the chosen values are $\phi_1 = 6$ and $\phi_2 = 1$.

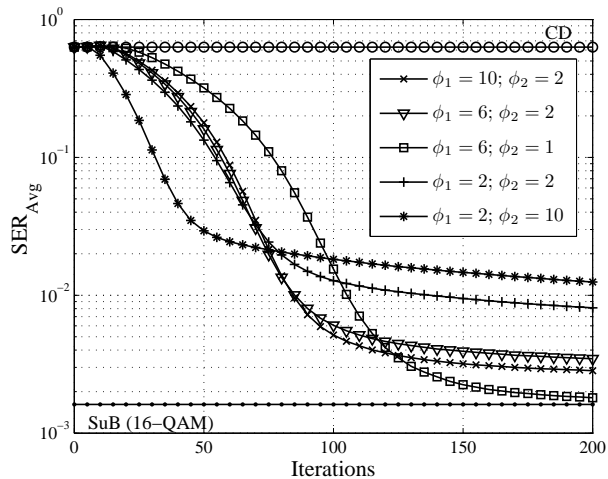

Fig. 8. $\phi_1$ and $\phi_2$ optimization under flat Rayleigh channels for 16-QAM modulation, $E_b/N_0 = 30$ dB, $K = 15$, $\omega = 1$ and $V_{\max} = 4$.

### C. Diversity Exploration

The best range for the acceleration coefficients under resolvable multipath channels ($L \geq 2$) for MuD SISO DS-CDMA problem seems $\phi_1 = 2$ and $\phi_2 \in [12; 15]$, as indicated by the simulation results shown in Fig. 9. For medium system loading and SNR, Fig. 9 indicates that the best values for acceleration coefficients are $\phi_1 = 2$ and $\phi_2 = 15$, allowing the combination of fast convergence and near-optimum performance achievement.

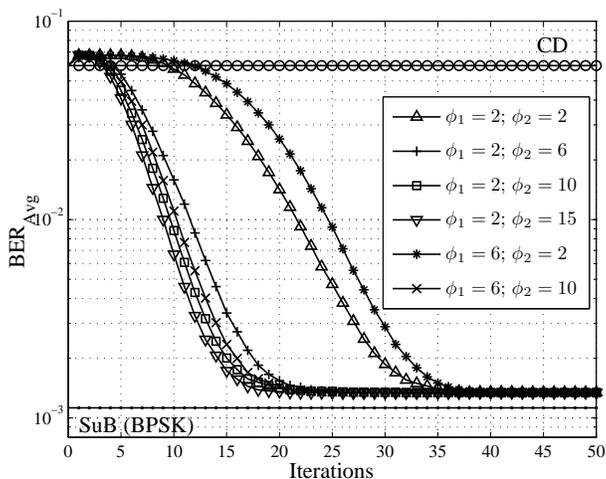

Fig. 9. $\phi_1$ and $\phi_2$ optimization under Rayleigh channels with path diversity ($L = D = 2$) for BPSK modulation, $E_b/N_0 = 22$ dB, $K = 15$, $\omega = 1$, $V_{\max} = 4$.

### D. Optimized parameters for PSO-MUD

As previously mentioned, the optimized input parameters for PSO-MUD vary regarding the system and channel scenario conditions. Monte-Carlo simulations exhibited in Section IV adopt the values presented in Tab. II as the optimized input PSO parameters. Loading system $\mathcal{L}$ range indicates the boundaries for

TABLE II

OPTIMIZED PARAMETERS FOR ASYNCHRONOUS PSO-MUD.

| Channel & Modulation | $\mathcal{L}$ range | $\omega$ | $\phi_1$ | $\phi_2$ | $V_{\max}$ |
|---|---|---|---|---|---|
| Flat Rayleigh BPSK | [0.16; 1.00] | 1 | 2 | 10 | 4 |
| Flat Rayleigh QPSK | [0.16; 1.00] | 1 | 4 | 4 | 4 |
| Flat Rayleigh 16-QAM | [0.03; 0.50] | 1 | 6 | 1 | 4 |
| Diversity Rayleigh BPSK | [0.03; 0.50] | 1 | 2 | 15 | 4 |

$\frac{K}{N}$ which the input PSO parameters optimization was carried out. For system operation characterized by spatial diversity ($Q > 1$ receive antennas), the PSO-MUD behaviour, in terms of convergence speed and quality of solution, is very similar to that presented under multipath diversity.

## IV. NUMERICAL RESULTS WITH OPTIMIZED PARAMETERS

In this section, numerical performance results are obtained using Monte-Carlo simulations. The results are compared with theoretical single-user bound (SuB), according to Appendix A, since the OMUD computational complexity results prohibitive. The adopted PSO-MUD parameters, as well as system and channel conditions employed in Monte Carlo simulations are summarized in Tab. III.

TABLE III

SYSTEM, CHANNEL AND PSO-MUD PARAMETERS FOR FADING CHANNELS PERFORMANCE ANALYSIS.

| Parameter | Adopted Values |
|---|---|
| *DS-CDMA System* | |
| # Rx antennas | $Q = 1, 2, 3$ |
| Spreading Sequences | Random, $N = 31$ |
| modulation | BPSK, QPSK and $16-$QAM |
| # mobile users | $K \in [5; 31]$ |
| Received SNR | $E_b/N_0 \in [0; 30]$ dB |
| *PSO-MUD Parameters* | |
| Population size, $\mathcal{P}$ | Eq. (21) |
| acceleration coefficients | $\phi_1 = 2, 6; \phi_2 = 1, 10$ |
| inertia weight | $\omega = 1$ |
| Maximal velocity | $V_{\max} = 4$ |
| *Rayleigh Channel* | |
| Channel state info. (CSI) | perfectly known at Rx |
| | coefficient error estimates |
| Number of paths | $L = 1, 2, 3$ |

Fig. 10 presents the performance as a function of received $E_b/N_0$ for two different near-far ratio scenarios under flat Rayleigh channel. Fig. 10.(a) was obtained for perfect power control, whereas Fig. 10.(b) was generated considering half users with $NFR = +6$ dB. Here, the BER$_{\text{Avg}}$ performance is calculated only for the weaker users. Note the performance of the PSO-MuD is almost constant despite of the $NFR = +6$ dB for half of the users, illustrating the robustness of the PSO-MuD against unbalanced powers in flat fading channels.

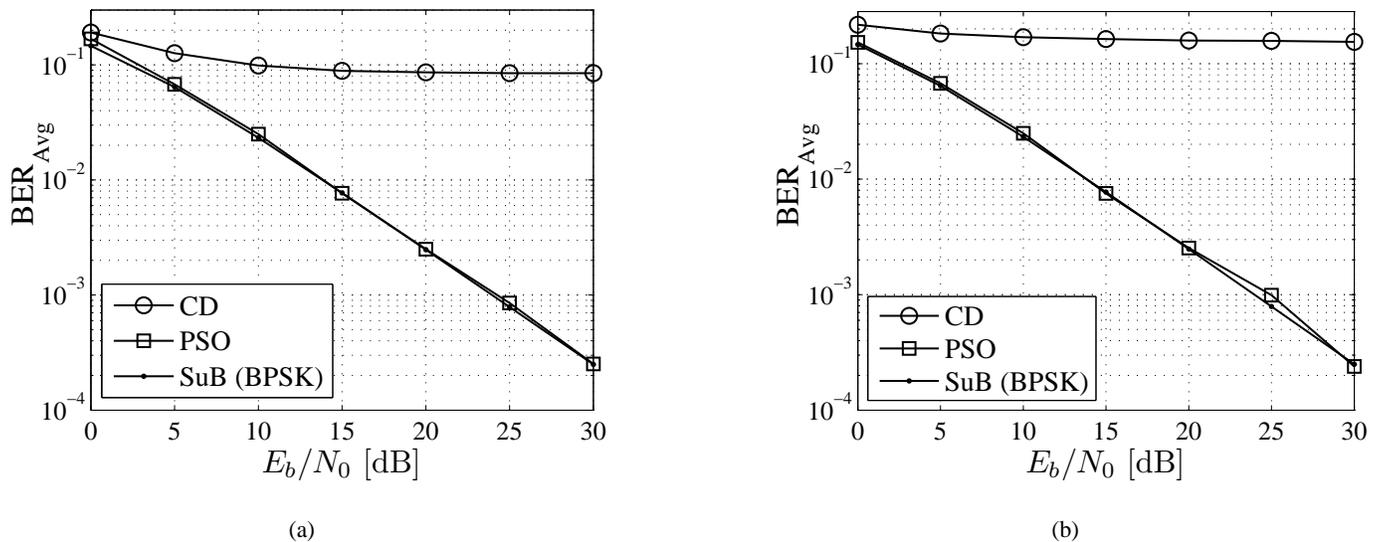

(a)         (b)

Fig. 10. Average BER$_{\text{Avg}} \times E_b/N_0$ for flat Rayleigh channel with $K = 15$: (a) perfect power control; (b) $NFR = +6$ dB for 7 users. In scenario (b), the performance is calculated only for the weaker users.

*A. Diversity*

In the results presented here, two assumptions are considered when there are more than one antenna at receiver (Spatial Diversity): first, the average received power is equal for all antennas; and second, the SNR at the receiver input is defined as the received SNR per antenna. Therefore, there is a power gain of 3 dB when adopted $Q = 2$, 6 dB with $Q = 3$, and so on. The effect of increasing the number of receive antennas in the convergence curves is shown in Fig. 11, where PSO-MuD works on systems with $Q = 1$, 2 and 3 antennas. A delay in the PSO-MuD convergence is observed when more antennas are added to the receiver, caused by the larger gap that it has to surpass. Furthermore, PSO-MuD achieves the SuB performance for all the three cases.

The exploitation of the path diversity also improves the system capacity. Fig. 11 shows the BER$_{\text{Avg}}$ convergence of PSO-MuD for different of paths, $L = 1$ 2 and 3, when the detector explores fully the path diversity, i.e., the number of fingers of conventional detector is equal the number of copies of signal received, $D = L$. The power delay profile considered is exponential, with mean paths energy as shown in Tab. IV [20]. It is worth mentioning that the mean received energy is equal for the three conditions, i.e., the resultant improvement with increasing number of paths is due the diversity gain only.

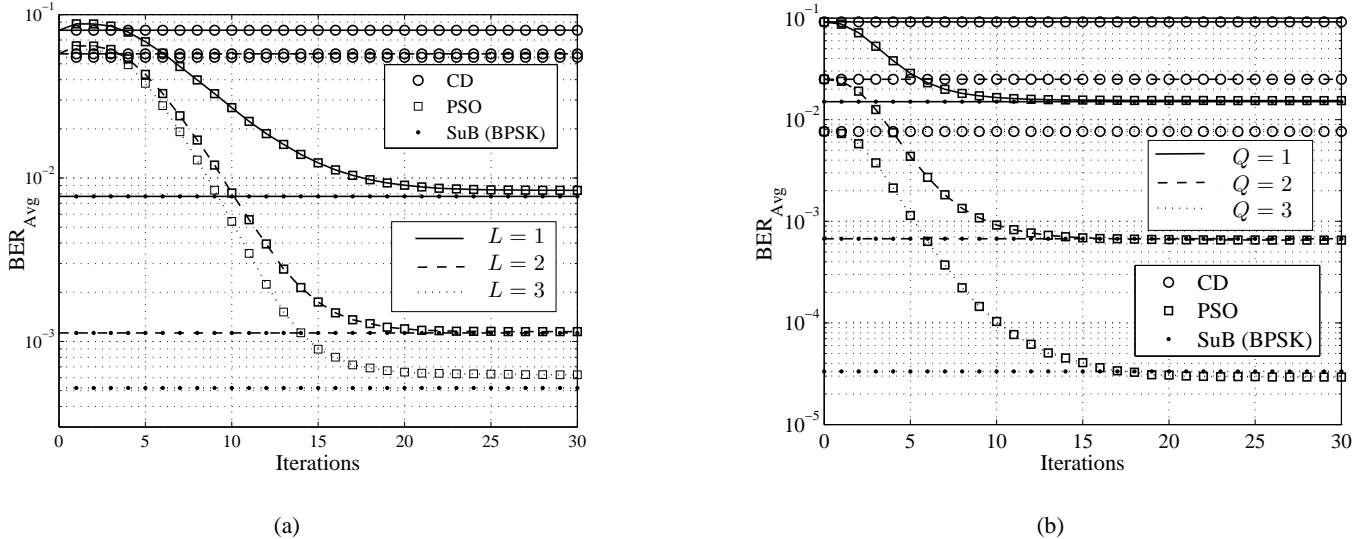

(a)

(b)

Fig. 11. Convergence performance of PSO-MUD, with $K = 15$, $E_b/N_0 = 15$, BPSK modulation, (a) under asynchronous multipath slow Rayleigh channels and $I = 3$, for $L = 1$, 2 and 3 paths; and (b) synchronous flat Rayleigh channel, $I = 1$ and $Q = 1, 2, 3$ antennas.

TABLE IV

THREE POWER-DELAY PROFILES FOR DIFFERENT RAYLEIGH FADING CHANNELS USED IN MONTE-CARLO SIMULATIONS.

| Param. | PD-1 | PD-2 | | PD-3 | | |
|---|---|---|---|---|---|---|
| Path, $\ell$ | 1 | 1 | 2 | 1 | 2 | 3 |
| $\tau_\ell$ | 0 | 0 | $T_c$ | 0 | $T_c$ | $2T_c$ |
| $\mathbb{E}[\gamma_\ell^2]$ | 1.0000 | 0.8320 | 0.1680 | 0.8047 | 0.1625 | 0.0328 |

Note there is a performance gain with the exploration of such diversity, verified in both the Rake receiver and PSO-MUD. The PSO-MUD performance is close to SuB in all cases, exhibiting its capability of exploring path diversity and dealing with SI as well. In addition, the convergence aspects are kept for all conditions.

The PSO-MUD is also evaluated under channel error estimation, which are modeled through the continuous uniform distributions $\mathcal{U}[1 \pm \epsilon]$ centralized on the true values of the coefficients, resulting

$$\widehat{\gamma}_{k,\ell}^{(i)} = \mathcal{U}[1 \pm \epsilon_\gamma] \times \gamma_{k,\ell}^{(i)}; \quad \widehat{\theta}_{k,\ell}^{(i)} = \mathcal{U}[1 \pm \epsilon_\theta] \times \theta_{k,\ell}^{(i)}, \qquad (22)$$

where $\epsilon_\gamma$ and $\epsilon_\theta$ are the maximum module and phase normalized errors for the channel coefficients, respectively. For a low-moderate SNR and medium system loading ($\mathcal{L} = 15/31$), Fig. 12 shows the performance degradation of the PSO-MUD considering BPSK modulation, $L = 1$ and $L = 2$ paths or $Q = 1$ and $Q = 2$ antennas, with estimation errors of order of 10% or 25%, i.e., $\epsilon_\gamma = \epsilon_\theta = 0.10$ or $\epsilon_\gamma = \epsilon_\theta = 0.25$, respectively. Note that PSO-MUD reaches the SuB in both conditions with perfect channel estimation, and the improvement is more evident when the diversity gain increases. However, note that, with spatial diversity, the gain is higher, since the average energy is equally distributed among antennas, while for path diversity is considered a realistic exponential power-delay profile. Although there is a general performance degradation when the error in channel coefficient estimation increases, PSO-MUD still achieves much better performance than the CD under any error estimation condition, being more evident for larger number of antennas.

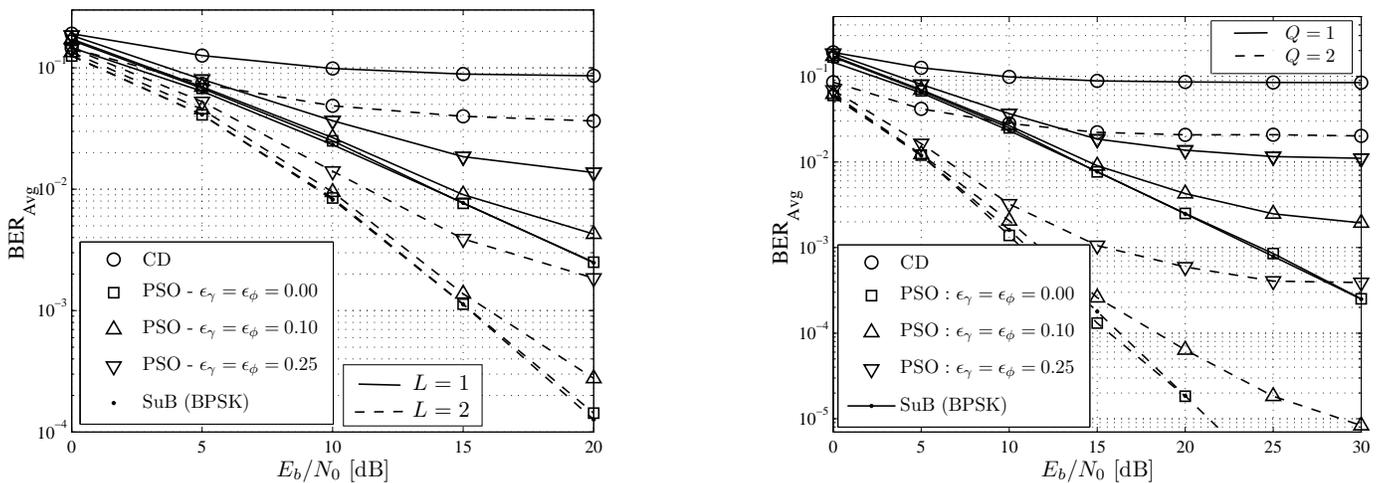

Fig. 12. Performance of PSO-MUD with $K = 15$, BPSK modulation and error in the channel estimation, for (a) path diversity and (b) spatial diversity.

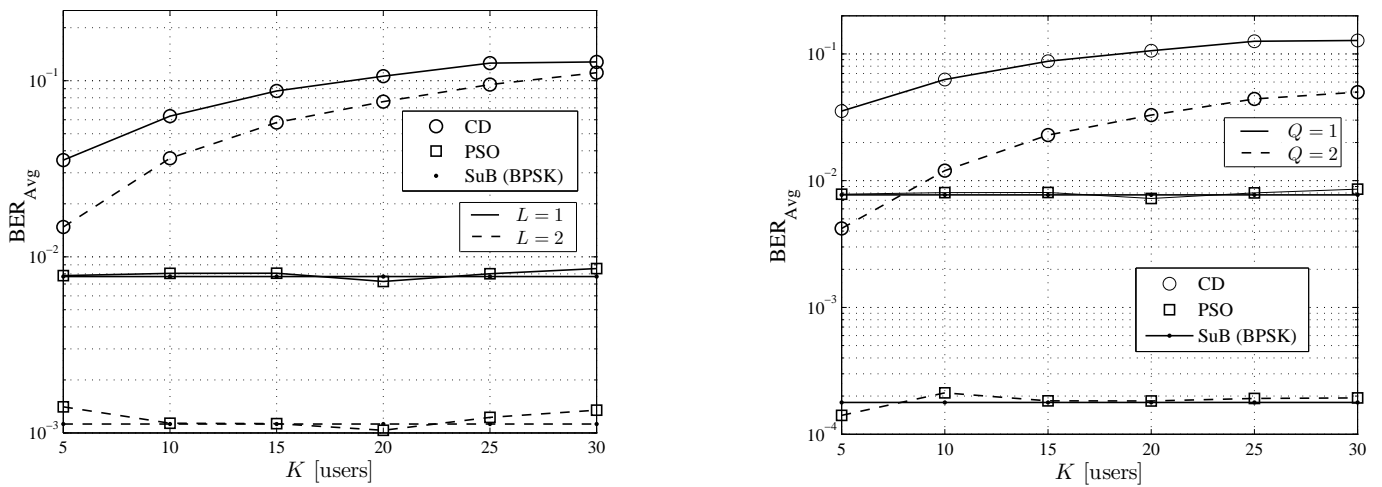

Fig. 13. Performance of PSO-MUD with $E_b/N_0 = 15$ dB and BPSK modulation, for (a) path diversity and (b) spatial diversity.

Fig. 13 shows the performance as function of number of users $K$. It is evident that the PSO-MUD performance is much superior then the CD scheme.

## B. QPSK and 16-QAM Modulations

Fig. 14 shows convergence comparison for three different modulations: (a) BPSK, (b) QPSK, and (c) 16-QAM. It is worth mentioning, as presented in Tab. II, that the PSO-MUD optimized parameters is specific for each modulation.

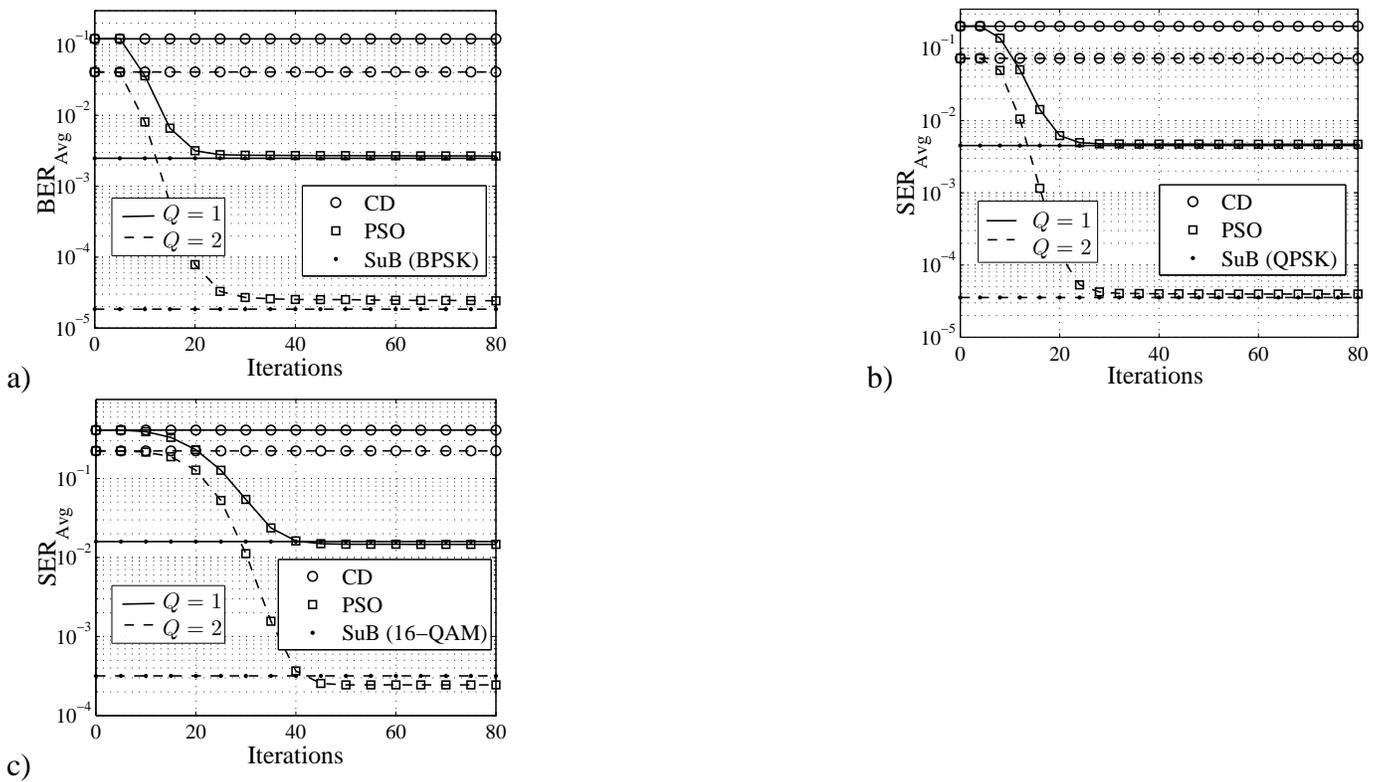

Fig. 14. Convergence of PSO-MUD under flat Rayleigh channel, $E_b/N_0 = 20$ dB, and a) $K = 24$ users with BPSK modulation, b) $K = 12$ users with QPSK modulation and c) $K = 6$ users with 16-QAM modulation

Similar results are obtained for $E_b/N_0$ curves in QPSK and 16-QAM cases. Nevertheless, Fig. 15 shows that for 16-QAM modulation with $\phi_1 = 6$, $\phi_2 = 1$, the PSO-MUD performance degradation is quite slight in the range $(0 < \mathcal{L} \leq 0.5)$, but the performance is hardly degraded in medium to high loading scenarios.

## V. CONCLUSION

This paper provides an analysis of the PSO scheme applied to the multiuser DS-CDMA system, focusing on the parameters optimization of the algorithm. It was shown that $\omega = 1$ represents a good choice for the considered detection problem and configurations.

Regarding the acceleration coefficients ($\phi_1$ and $phi_2$) in Rayleigh flat channels, it was demonstrated that their choices depend on the modulation order. With BPSK $\phi_1 = 2$ and $\phi_2 = 10$ represent a good choice. For QPSK $\phi_1 = 4$ and $\phi_2 = 4$ represented a good complexity $\times$ performance trade-off, while for 16-QAM, it was observed that $\phi_1 = 6$ and $\phi_2 = 1$ provide a good result. However, in the latter case, the performance is not optimum for high system loading. With BPSK and Rayleigh diversity channels, it was shown that $\phi_1 = 2$ and $\phi_2 = 12$ to $15$ provide a good convergence of the PSO.

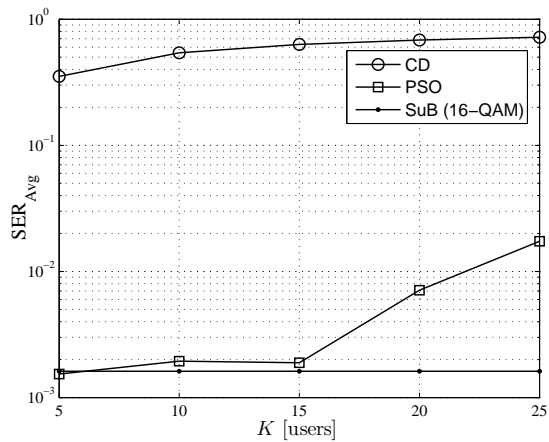

Fig. 15. PSO-MUD and CD performance degradation × system loading under 16-QAM modulation, in flat Rayleigh channel.

The PSO algorithm shows to be efficient for SISO/SIMO MUD asynchronous DS-CDMA problem when the input parameters are properly chosen. Under a variety of simulated/analyzed realistic scenarios, the performance achieved by PSO-MUD, except for high order modulation in the high system loading condition, was near-optimal. In the presence of channel errors, the PSO-MUD keeps much more efficient than conventional receiver with perfect channel estimation. In all evaluated system conditions, PSO-MUD resulted in small degradation performance if those errors are confined to 10% of the actual instantaneous values.

## APPENDIX

### A. Minimal Number of Trials and Single-User Performance

The minimal number of trials ($TR$) evaluated in the each simulated point (SNR) was obtained based on the single-user bound (SuB) performance. Considering a confidence interval, and admitting that a non-spreading and a spreading systems have the same equivalent bandwidth ($BW \approx \frac{1}{T_s} = BW_{\text{spread}} \approx \frac{N}{T_c}$), and thus, equivalently, both systems have the same channel response (delay spread, diversity order and so on), the SuB performance in both systems will be equivalent. So, the average symbol error rate for a single-user under $M$-QAM DS-CDMA system and $L$ Rayleigh fading path channels with exponential power-delay profile and maximum ratio combining reception is found in [22, Eq. (9.26)] as

$$\text{SER}_{\text{SuB}} = 2\alpha \sum_{\ell=1}^{L} p_\ell(1-\beta_\ell) + \tag{23}$$

$$\alpha^2 \left[ \frac{4}{\pi} \sum_{\ell=1}^{L} p_\ell \beta_\ell \times \tan^{-1}\left(\frac{1}{\beta_\ell}\right) - \sum_{\ell=1}^{L} p_\ell \right]$$

where:

$$p_\ell = \left( \prod_{k=1, k\neq\ell}^{L} \left(1 - \frac{\overline{\nu}_k}{\overline{\nu}_\ell}\right) \right)^{-1}, \quad \alpha = \left(1 - \frac{1}{\sqrt{M}}\right),$$

$$\beta_\ell = \sqrt{\frac{\overline{\nu}_\ell g_{\text{QAM}}}{1 + \overline{\nu}_\ell g_{\text{QAM}}}}, \quad g_{\text{QAM}} = \frac{3}{2(M-1)},$$

and $\overline{\nu}_\ell = \overline{\nu}_\ell^* \log_2 M = m\overline{\nu}_\ell^*$ denotes the average received signal-noise ratio per symbol for the $\ell$th path, with $\overline{\nu}_\ell^*$ being the correspondent SNR per bit per path.

Once the lower bound is defined, the minimal number of trials can be defined as

$$\text{TR} = \frac{n_{\text{errors}}}{\text{SER}_{\text{SuB}}},$$

where the higher $n_{\text{errors}}$ value, the more reliable will be the estimate of the SER obtained in MCS [8]. In this work, the minimum adopted $n_{\text{errors}} = 100$, and considering a reliable interval of $95\%$, it is assured that the estimate $\widehat{\text{SER}} \subset [0.823;\, 1.215]\,\text{SER}$. Simulations were carried out using MATLAB v.7.3 plataform, The MathWorks, Inc.